\documentclass[11pt,letterpaper]{Sengoku}

\usepackage{amsmath,amsfonts,bm}

\def\eqref#1{equation~\ref{#1}}
\def\1{\bm{1}}

\DeclareMathAlphabet{\mathsfit}{\encodingdefault}{\sfdefault}{m}{sl}
\SetMathAlphabet{\mathsfit}{bold}{\encodingdefault}{\sfdefault}{bx}{n}

\usepackage{booktabs}
\usepackage{multirow}
\usepackage{wrapfig}
\usepackage{float}

\title{\gradienttitle{DK-VLM:}{Geometry as Position}{for Spatial Reasoning}{in Vision-Language Models}}
\date{\today}
\shorttitle{DK-VLM: Geometry as Position}
\toplogogap{5mm}
\addtoplogo[height=10mm]{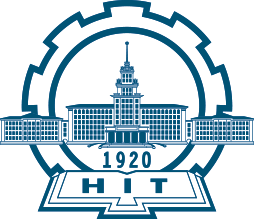}
\addtoplogo[height=6mm]{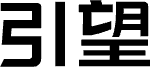}
\addtoplogo[height=6mm]{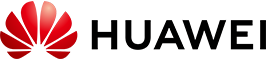}
\ExplSyntaxOn
\cs_new_eq:NN \DKVLMOriginalTopLogoBlock \sengoku_render_top_logo_block:
\ExplSyntaxOff
\RenewDocumentCommand{\SengokuRenderTopLogoBlock}{}{%
  \DKVLMOriginalTopLogoBlock
  \noindent\hspace*{.03\textwidth}{\color{SengokuAccent}\rule{.94\textwidth}{0.45pt}}\par
  \vspace{2mm}%
}

\affiliationlayout{inline}
\affiliation{hit}{Harbin Institute of Technology}
\affiliation{yinwang}{Yinwang Intelligent Technology}
\affiliation{huawei}{Huawei}

\author[hit,yinwang]{Han Su}
\author[yinwang]{Meng Tian}
\author[huawei]{Tianyu Huang}
\author[hit]{Mingyu Gao}
\author[hit]{Lennox Wei}
\author[hit]{Dongxu Li}
\author[yinwang]{Jianhua Han}
\author[yinwang]{Hang Xu}
\author[hit]{Xiaohe Wu}
\author[]{Wangmeng Zuo\textsuperscript{1,*}}

\hypersetup{pdfsubject={Spatial reasoning in vision-language models},pdfauthor={Han Su; Meng Tian; Tianyu Huang; Mingyu Gao; Lennox Wei; Dongxu Li; Jianhua Han; Hang Xu; Xiaohe Wu; Wangmeng Zuo}}
\makeatletter
\gdef\Sengoku@titlenotes{\centering\textsuperscript{*} Corresponding author.\par}
\makeatother

\begin{document}
\vspace*{-13mm}

\begin{abstract}
Vision-language models still struggle to understand 3D spatial relations across views and update spatial judgments as the observer moves.
A common approach fuses 3D reconstruction features into visual tokens, but aligning these features with pretrained semantic representations remains challenging under limited spatial supervision.
We propose DK-VLM with \emph{Dual-Kernel Attention}, which computes attention from both a visual token's input position and its geometric position in the scene.
A semantic kernel retains the pretrained model's native attention and positional encoding, while a separate geometric kernel uses estimated 3D points and camera poses to directly modulate attention through point displacements and relative view transformations.
Then, a learned gate fuses their outputs.
Geometry thus enters attention as position rather than as fused reconstruction features.
We further introduce Reference-Frame Composition (RC) supervision to train the model to update its reference frame across successive judgments.
It provides stepwise turn and heading targets derived from verified single-step annotations.
Among the compared methods, DK-VLM-8B achieves the highest average score across six spatial benchmarks and the highest overall scores on VSI-Bench (74.33) and ReVSI-32 (58.28).
RC supervision further improves multi-step spatial reasoning, while the full model preserves general video understanding.
Project page: \url{https://sumuru789.github.io/DK-VLM-website/}.
\end{abstract}

\maketitle
\enlargethispage{.5\baselineskip}

\begin{figure}[!t]
    \centering
    \includegraphics[width=\textwidth]{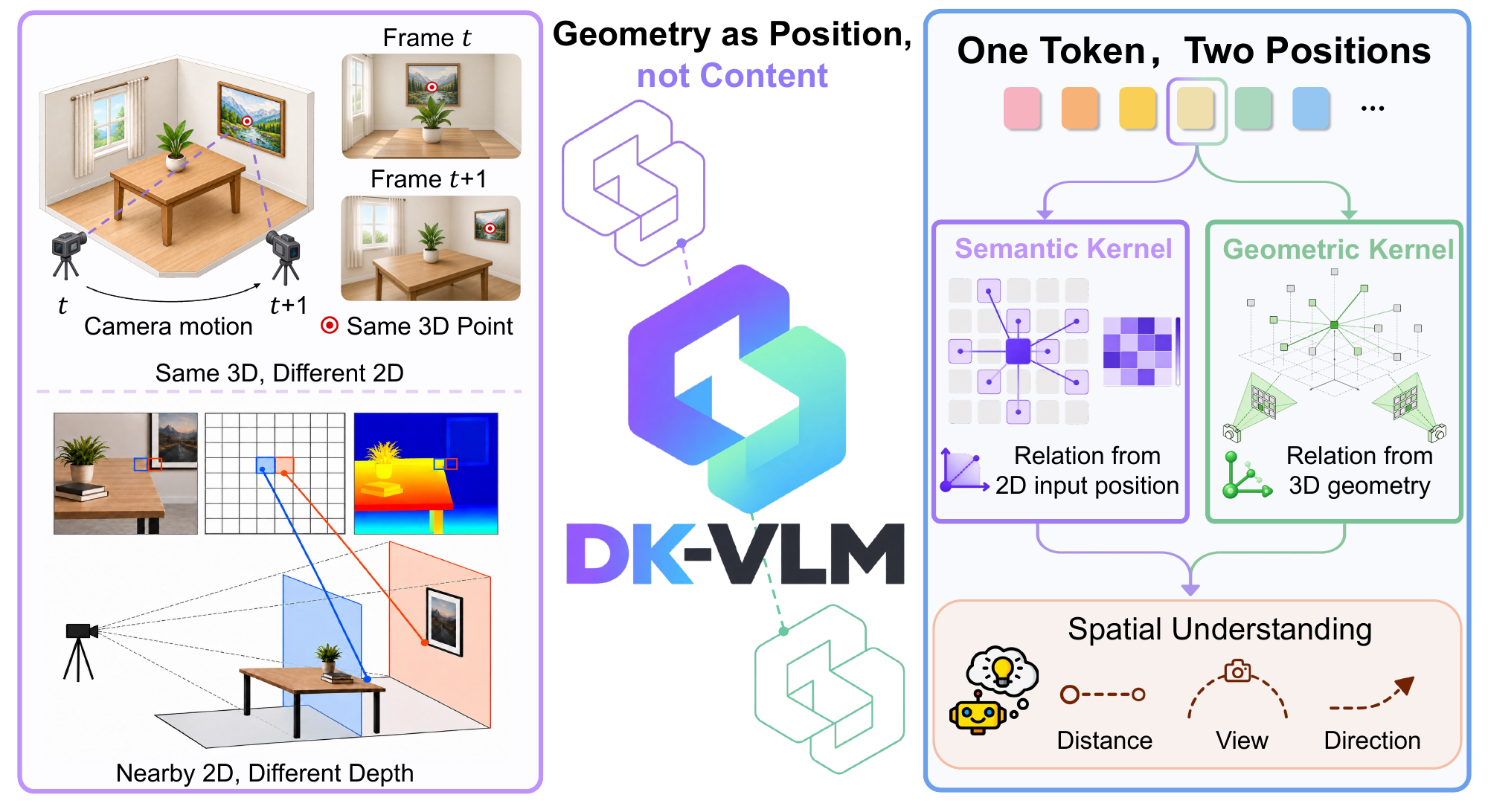}
    \caption{\textbf{Geometry as position in DK-VLM.}
    Left: 2D input positions do not fully capture 3D relations. Right: DK-VLM retains the native positional encoding and introduces scene geometry into attention as position.}
    \label{fig:teaser}
\end{figure}

\section{Introduction}
\label{sec:introduction}

Spatial reasoning connects visual understanding to agentic action in the physical world~\citep{chen2024spatialvlm,su2026s2am3d,huang2025dreamphysics,zhang2025physchoreo,driess2023palme,huang2023voxposer}. Tasks like multi-view question answering, long-video understanding, robot manipulation, and embodied planning require more than recognizing visual content~\citep{yang2025vsibench,mangalam2023egoschema}.
Specifically, a vision-language model (VLM) must also perceive 3D relations across views and, as the observer moves, keep updating the position and heading on which its subsequent spatial judgments depend~\citep{chen2024spatialvlm,yang2025vsibench}. Despite strong progress in semantic recognition and language reasoning, VLMs remain unreliable at estimating distance, direction, camera motion, and cross-view correspondence, and weaker still on tasks such as route planning that chain several such judgments~\citep{yang2025vsibench,zhang2026revsi}.

Existing work attempts to enhance spatial reasoning by introducing 3D priors. One common approach fuses features from pretrained 3D reconstruction models with visual representations, incorporating geometry into the content of visual tokens~\citep{fan2026vlm3r,li2026geothinker}. However, these geometric features are primarily learned for 3D reconstruction, an objective different from semantic understanding in VLMs. Integrating them therefore requires the model to align two different representation spaces while learning spatial relations. With limited spatial supervision, this additional adaptation may limit the effective use of geometry and interfere with the model's existing semantic representations.

Beyond token content, VLMs already use positional encoding to guide attention. In the Qwen-VL family~\citep{wang2024qwen2vl,bai2025qwen25vl,bai2025qwen3vl}, multi-modal rotary position embedding (mRoPE) encodes sequence order, time, and 2D image grid coordinates, so that attention between two tokens depends on their relative positions in the input. Position thus offers a second interface for geometry: instead of being fused into token content, geometry can define relations between tokens that directly shape attention~\citep{miyato2024gta,li2025prope,wu2026rayrope,xiang2026viewrope}. However, input positions describe where a token appears in the input rather than where its content lies in the 3D world. As the camera moves, the same 3D surface appears at different image locations across frames, while neighboring 2D patches can lie at very different depths (Figure~\ref{fig:teaser}).

Existing geometric attention mechanisms~\citep{miyato2024gta,li2025prope,wu2026rayrope,xiang2026viewrope} introduce 3D relations directly into attention by applying geometric transformations to queries, keys, or values. However, replacing a pretrained VLM's native positional encoding with geometric encoding may disrupt the semantic attention patterns learned under the original encoding---patterns that remain useful for spatial understanding.
We therefore bring scene geometry into attention through a separate kernel while retaining the native attention and positional encoding.

We introduce DK-VLM, which implements this design through \emph{Dual-Kernel Attention}: attention is computed from both a visual token's input position and its geometric position in the scene. The semantic kernel keeps the backbone's native attention and mRoPE unchanged. The geometric kernel uses independent projections and computes relations between visual tokens conditioned on patch-level 3D points and frame-level camera poses, so that displacements between scene points and relative transformations between views directly modulate feature comparison and aggregation. A learned gate fuses the two outputs.
Geometry thus enters attention as position rather than as reconstruction features fused into token content. The geometric inputs are estimated from images, requiring no ground-truth depth or camera poses at inference.

Dual-Kernel Attention gives the model direct access to point displacements and relative view transformations, but successive spatial judgments require more: earlier movements change the position and heading on which later judgments depend, even when the scene layout remains unchanged. Single-step question answering supervises each judgment within a given reference frame and does not explicitly supervise how that frame changes across steps. We therefore introduce Reference-Frame Composition (RC) supervision, which provides stepwise turn and heading targets along landmark routes and view sequences. These targets are derived by connecting verified single-step relations that share landmarks or views within the same scene~\citep{yang2025cambrians,zhang2025spar}.

We implement DK-VLM on Qwen2.5-VL-7B and Qwen3-VL-8B~\citep{bai2025qwen25vl,bai2025qwen3vl}. We evaluate it on seven spatial reasoning benchmarks~\citep{yang2025vsibench,zhang2026revsi,yang2025mmsibench,yin2025mindcube,li2025viewspatial,wang2025site,tong2024cambrian1} and two general video understanding benchmarks~\citep{fu2025videomme,li2024mvbench}. Among the compared methods, DK-VLM-8B achieves the highest six-benchmark average score and the highest overall scores on VSI-Bench (74.33) and ReVSI-32 (58.28), with trainable parameters totaling only 3.86\% of its backbone's parameter count. RC supervision further improves multi-step spatial reasoning, yielding consistent gains on route-planning and viewpoint-related tasks. Ablations show that DK-VLM benefits more from RC supervision than a Qwen baseline fine-tuned on the same instruction data, supporting the complementarity between geometric relation modeling and stepwise spatial supervision. The full model also preserves general video understanding on VideoMME and MVBench.

Our contributions are:
\begin{itemize}
    \item \textbf{Dual-Kernel Attention}, which introduces scene geometry as position through a geometric kernel parallel to a pretrained VLM's native attention.
    \item \textbf{Reference-Frame Composition supervision}, which supervises reference-frame updates using stepwise targets derived from verified single-step relations.
    \item DK-VLM-8B achieves the highest overall scores among compared methods on VSI-Bench and ReVSI-32 and the highest average in the six-benchmark comparison.
\end{itemize}

\section{Related Work}
\label{sec:related_work}

\paragraph{Learning Spatial Reasoning.}
Recent work improves spatial reasoning in VLMs through spatial data and reasoning-oriented training. VST and Cambrian-S scale spatial instruction tuning, SpatialLadder adopts a progressive curriculum from perception to reasoning, and SpaceR and SpatialEvo respectively introduce verifiable rewards and deterministic geometric environments~\citep{yang2025vst,yang2025cambrians,li2025spatialladder,ouyang2025spacer,li2026spatialevo}. However, compositional supervision for tracking reference frame changes across observations has received less attention.

\paragraph{Geometry-Augmented Vision-Language Models.}
% Previous opening (before 2026-09-22; coordinate-injection works were cited only in the introduction): Another line of work extracts cross-view geometry from 3D reconstruction models and integrates it into VLMs.
% Previous version (before 2026-09-22): Another line of work introduces 3D information into VLMs. Video-3D LLM, LLaVA-3D, and HiSpatial encode 3D coordinates into visual tokens, while other methods extract cross-view geometry from 3D reconstruction models and integrate it into VLMs. VG-LLM, VLM-3R, and GeoThinker follow this route, with additional discussion of camera guidance, hierarchical injection, and G$^2$VLM.
One line of work introduces 3D information by encoding coordinates into visual tokens, including Video-3D LLM, LLaVA-3D, and HiSpatial~\citep{zheng2024video3dllm,zhu2024llava3d,liang2026hispatial}. Video-3D LLM adds coordinate embeddings to visual features. Another line introduces features from 3D reconstruction models: VG-LLM adds projected VGGT features to visual tokens, VLM-3R queries reconstruction and camera tokens with cross-attention, and GeoThinker retrieves geometric features according to the current semantic state~\citep{zheng2025vgllm,fan2026vlm3r,li2026geothinker}. These fusion mechanisms let a VLM use geometric representations from reconstruction models, but require aligning features learned for reconstruction with pretrained semantic representations. In contrast, DK-VLM uses estimated 3D points and camera poses to directly modulate attention among visual tokens, rather than incorporating coordinate embeddings or reconstruction features into their content.

\paragraph{Geometric Attention and Positional Encoding.}
Positional mechanisms provide another interface for introducing 3D information into Transformers. Geometric Transform Attention (GTA) incorporates group actions induced by camera transformations into attention, while PRoPE, RayRoPE, and ViewRoPE use 3D points, rays, or viewpoints to parameterize positional relations between tokens~\citep{miyato2024gta,li2025prope,wu2026rayrope,xiang2026viewrope}. These studies show that geometry can directly shape relation computation. How to introduce these mechanisms into a pretrained VLM while retaining its native positional encoding remains less explored.

\section{Method}
\label{sec:method}

\begin{figure*}[t]
    \centering
    \includegraphics[width=\textwidth]{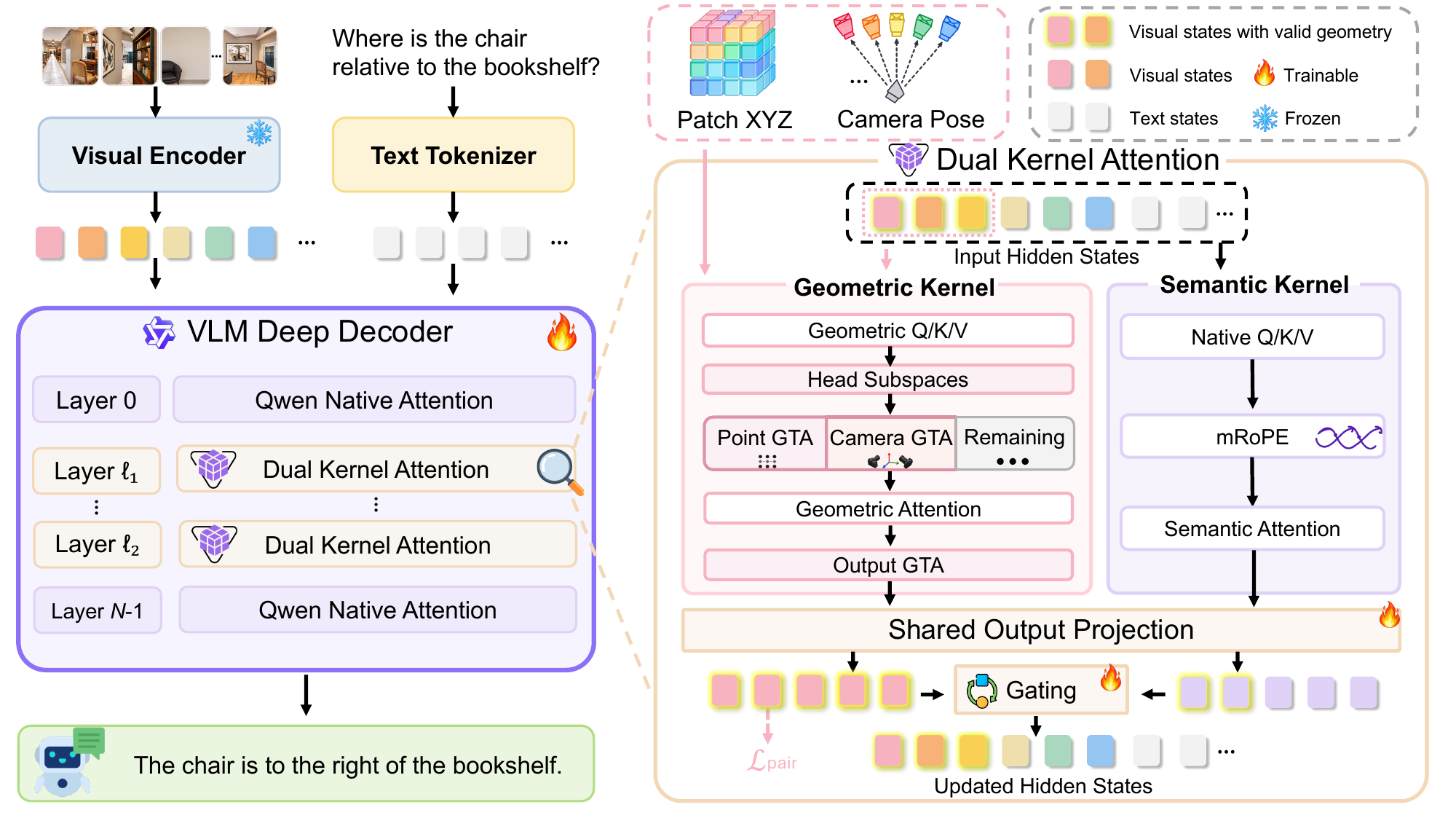}
    \caption{
    \textbf{Geometry as Position via Dual-Kernel Attention.}
    Selected decoder layers retain the backbone-native semantic kernel and introduce a geometric kernel over geometry-valid visual tokens. A learned gate fuses both outputs into the shared residual stream, while $\mathcal L_{\mathrm{pair}}$ directly supervises the pre-fusion geometric representation during training.
    }
    \label{fig:method}
\end{figure*}

Given a sequence of RGB images and a spatial query, DK-VLM uses a geometry frontend to estimate aligned patch-level 3D points and frame-level camera poses from the images~\citep{wang2025vggt,wang2026vggtomega,wang2026pi3,lin2025da3,murai2025mast3rslam}. DK-VLM uses these geometric states rather than latent features specific to a frontend, allowing compatible geometry frontends to be replaced. It models input position and scene geometry through parallel relation kernels. The semantic kernel retains the backbone's native positional computation, while the geometric kernel is parameterized by the estimated states. Dual-Kernel Attention computes and fuses the outputs of the two kernels (Section~\ref{sec:dual-kernel}). To supervise the composition of geometric relations across observations, Reference-Frame Composition links verified single-step annotations into training chains with stepwise targets (Section~\ref{sec:reference-frame-composition}). Training jointly optimizes the instruction and pairwise geometric objectives in a single stage (Section~\ref{sec:joint-training-objective}). Figure~\ref{fig:method} illustrates the dual-kernel architecture.

\subsection{Dual-Kernel Attention}
\label{sec:dual-kernel}

\paragraph{Dual-Kernel Formulation.}
To introduce 3D relations while retaining the backbone-native attention computation, we use two relation kernels in parallel. The semantic kernel preserves the semantic, temporal, and two-dimensional image relations learned during pretraining, while the geometric kernel conditions relations among geometry-valid visual tokens on 3D points and camera poses. At each selected Transformer layer, let $H^{\ell}\in\mathbb{R}^{L\times d}$ be the layer input, with sequence length $L$ and hidden dimension $d$. Both kernels read $H^{\ell}$ and produce head-space outputs $S^{\ell}$ and $G^{\ell}$. At other layers, attention follows the backbone's native computation.

Each visual token $i$ with valid geometry carries a 3D point $\mathbf{x}_i\in\mathbb{R}^3$ in a common scene coordinate system. It is also associated with the camera-to-world pose $C_i^{\mathrm{rel}}\in SE(3)$ of its source frame, expressed relative to the first frame. We denote its rotation and translation by $R_i^{\mathrm{rel}}$ and $\mathbf t_i^{\mathrm{rel}}$; same-frame tokens share these quantities. Let $\mathcal{J}$ index tokens with both states valid.

The semantic kernel retains the backbone-native Q/K/V projections, mRoPE, and causal and padding masks~\citep{wang2024qwen2vl,bai2025qwen25vl,bai2025qwen3vl}; $S^{\ell}$ is its output before the native output projection. Through independent projections, the geometric kernel derives $Q_g=H^{\ell}W^{\ell}_{Q,g}$, $K_g=H^{\ell}W^{\ell}_{K,g}$, and $V_g=H^{\ell}W^{\ell}_{V,g}$ from the same hidden states and conditions their interactions on aligned geometry. These Q/K/V projections are initialized from the corresponding native projections and have separate trainable parameters. Because $\mathcal J$ contains only observed visual tokens, a causal mask would make geometric relation computation depend on input order: earlier tokens could not use geometric evidence from observations appearing later in the sequence. We therefore apply full attention over $\mathcal J$, allowing each geometry-valid visual token to interact with all input frames; text and geometry-invalid tokens are excluded.

\paragraph{Geometric Kernel.}
We use two forms of GTA~\citep{miyato2024gta} to introduce multi-view geometry into attention: Point GTA for relations between scene points and Camera GTA for camera motion across views. We split each projected geometric Q/K/V head into point, camera, and remaining channels, with the two transforms acting on their respective subspaces. For queries, this gives $\mathbf q_i=[\mathbf q_i^{\mathrm{pt}}\Vert\mathbf q_i^{\mathrm{cam}}\Vert\mathbf q_i^{\mathrm{rem}}]$, with the same split for keys and values. The remaining channels keep their independently projected features and participate in the same geometric attention.

GTA incorporates relative geometric transforms into both attention weights and feature aggregation. Both forms use the same GTA operation: we divide the feature channels of the point and camera subspaces into blocks of four and apply the corresponding $4\times4$ transform to each block. Let $\mathbf q_{i,b},\mathbf k_{i,b},\mathbf v_{i,b},\mathbf o_{i,b}\in\mathbb R^4$ denote the query, key, value, and attention output features in block $b$ of token $i$. Given a transform $T_{i,b}\in SE(3)$ and its inverse $P_{i,b}=T_{i,b}^{-1}$, GTA transforms Q/K/V before attention and the output afterward:

\begin{equation}
\widehat{\mathbf q}_{i,b}=P_{i,b}^{\top}\mathbf q_{i,b},\qquad
\bigl[\widehat{\mathbf k}_{i,b}\;\widehat{\mathbf v}_{i,b}\bigr]
=T_{i,b}\bigl[\mathbf k_{i,b}\;\mathbf v_{i,b}\bigr],\qquad
\widehat{\mathbf o}_{i,b}=P_{i,b}\mathbf o_{i,b}.
\label{eq:gta-action}
\end{equation}

The transformed query-key inner product is $\widehat{\mathbf q}_{i,b}^{\top}\widehat{\mathbf k}_{j,b}=\mathbf q_{i,b}^{\top}T_{i,b}^{-1}T_{j,b}\mathbf k_{j,b}$. Thus, the geometric transforms applied to the two tokens combine into a relative transform in the attention score.

Camera GTA uses the camera pose of the source frame, with $T^{\mathrm{cam}}_{i,b}=C_i^{\mathrm{rel}}$ for all camera feature blocks. Tokens from the same frame share this transform, whose relative action $(C_i^{\mathrm{rel}})^{-1}C_j^{\mathrm{rel}}$ describes camera motion between views.

However, spatial understanding also requires 3D relations between scene points. Camera poses alone do not specify the scene position of each patch. We therefore construct Point GTA using the patch's scene point as the translation anchor, while keeping the source frame orientation. The transform for feature block $b$ is

\begin{equation}
T^{\mathrm{pt}}_{i,b}
=
\begin{bmatrix}
R_i^{\mathrm{rel}} & \lambda_b\mathbf{x}_i\\
\mathbf{0}^{\top} & 1
\end{bmatrix},
\qquad
P^{\mathrm{pt}}_{i,b}=\bigl(T^{\mathrm{pt}}_{i,b}\bigr)^{-1}.
\label{eq:point-gta}
\end{equation}

Here, $\lambda_b$ scales the displacement in feature block $b$. In the relative transform $\bigl(T^{\mathrm{pt}}_{i,b}\bigr)^{-1}T^{\mathrm{pt}}_{j,b}$, translation depends on the difference between point positions, while rotation captures the relative orientation of the source views:

\begin{equation}
R_{ij}^{\mathrm{view}}
= {R_i^{\mathrm{rel}}}^{\top}R_j^{\mathrm{rel}},
\qquad
\mathbf t_{ij,b}^{\mathrm{pt}}
= \lambda_b{R_i^{\mathrm{rel}}}^{\top}(\mathbf x_j-\mathbf x_i).
\label{eq:point-gta-relative-components}
\end{equation}

The full matrix derivation is given in Appendix~\ref{app:point-gta-expansion}.

The feature blocks transformed by Point GTA and Camera GTA, together with the remaining channels, form $\widetilde Q_g$, $\widetilde K_g$, and $\widetilde V_g$. We gather the visual tokens with valid geometry using the index set $\mathcal J$ and compute geometric attention:

\begin{equation}
O^{\ell}_{g,\mathcal{J}}
=
\operatorname{Softmax}\!\left(
\frac{\widetilde Q_{g,\mathcal{J}}
\widetilde K_{g,\mathcal{J}}^{\top}}
{\sqrt{d_h}}
\right)
\widetilde V_{g,\mathcal{J}},
\label{eq:geometric-attention}
\end{equation}

where $d_h$ is the head dimension. Applying the inverse GTA transform to $O^\ell_{g,\mathcal J}$ introduces relative token transforms into value aggregation as well, yielding $G^\ell_{\mathcal J}$. We then write these outputs back to their corresponding sequence positions to fuse the two kernels.

\paragraph{Kernel Fusion.}
Both kernels share an output projection $\phi_g^{\ell}$, initialized from the native output projection and trained independently of it. A separate scalar gate at each selected layer learns the mixing ratio between the two kernels. For positions $i\in\mathcal J$, we fuse their outputs as:

\begin{equation}
Y_i^{\ell}
=
(1-\alpha_{\ell})\,\phi_g^{\ell}(S_i^{\ell})
+
\alpha_{\ell}\,\phi_g^{\ell}(G_i^{\ell}),
\qquad
\alpha_{\ell}=\sigma(g_{\ell}).
\label{eq:kernel-fusion}
\end{equation}

We initialize $g_{\ell}=0$ ($\alpha_{\ell}=0.5$); training then learns an independent mixture for each selected layer. At positions outside $\mathcal J$, the semantic output passes through its native output projection. The fused outputs are written back to their original positions and passed to the residual stream.

\subsection{Reference-Frame Composition}
\label{sec:reference-frame-composition}

\begin{wrapfigure}{r}{0.44\textwidth}
    \vspace{-0.7\baselineskip}
    \centering
    \includegraphics[width=\linewidth]{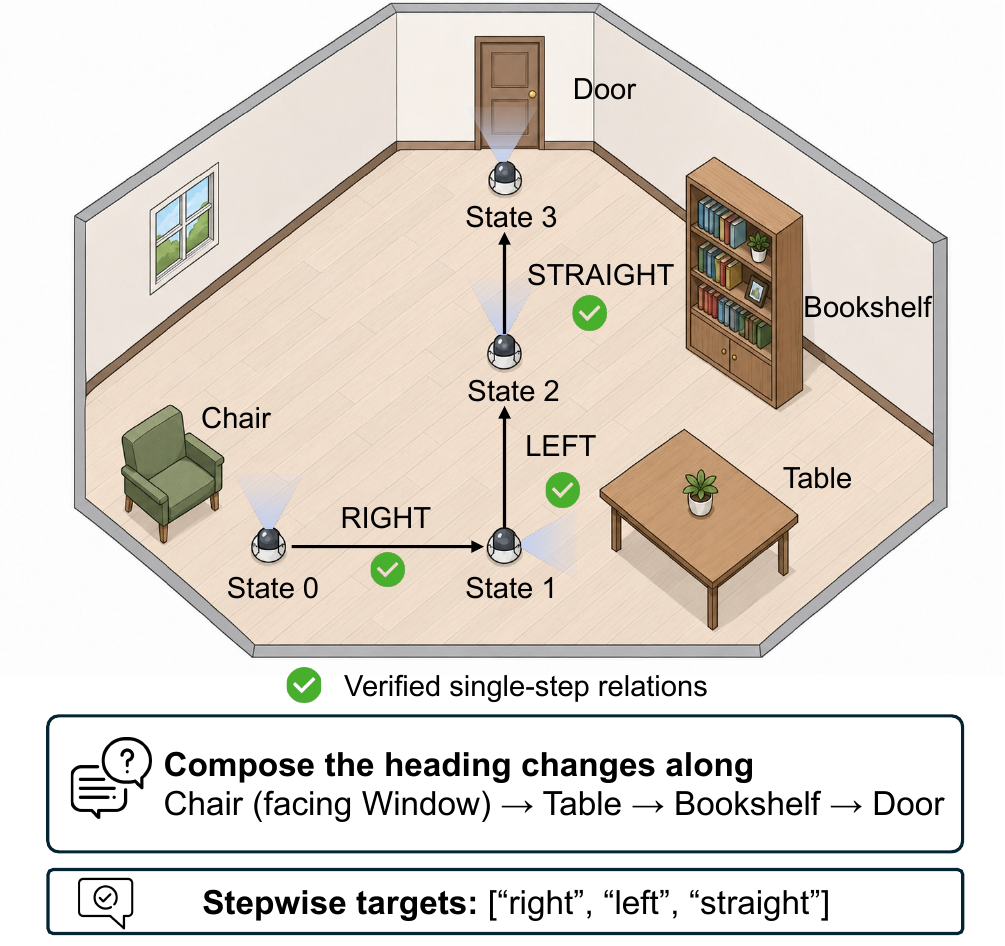}
    \caption{\textbf{Reference-frame composition.}
    Verified single-step relations sharing endpoints form a training chain with stepwise turn targets.}
    \label{fig:reference-frame-composition}
    \vspace{-1.0\baselineskip}
\end{wrapfigure}

Dual-Kernel Attention conditions token relations on 3D position and relative pose, but multi-step reference-frame reasoning additionally requires composing relative transformations across observations. Standard single-step supervision guides spatial judgments within a given reference frame, but does not explicitly supervise how that frame changes across successive steps. Reference-Frame Composition (RC) instead derives and supervises the reference-frame update at every step. Figure~\ref{fig:reference-frame-composition} illustrates one such chain and its stepwise targets.

\paragraph{Chain Construction.}
RC first converts each verified relation according to the direction convention of its source. Compatible relations are then connected along a sequence of landmarks or views to form a path $\mathcal C=(v_0,\ldots,v_K)$. The heading obtained at one step provides the reference for the next, turning the path into a sequence of reference-frame updates. The target at each step is recomputed from the source annotations; Appendix~\ref{app:reference-frame-composition} details the construction and verification.

\paragraph{Stepwise Supervision.}
Each retained relation is converted under its source coordinate convention into a canonical clockwise planar turn increment $\delta_k\in[0,360^\circ)$. RC describes the resulting reference-frame composition as

\begin{equation}
h_0=0,\qquad
h_k=(h_{k-1}+\delta_k)\bmod 360^\circ.
\label{eq:reference-frame-composition}
\end{equation}

Here $h_k$ is the reference-frame heading after transition $k$, for $1\leq k\leq K$. RC uses the turn increments and, when available, their cumulative headings as stepwise targets.

We apply RC to VSI-590K and the view-change subset of SPAR-7M to construct four types of supervision~\citep{yang2025cambrians,zhang2025spar}: exact turn angles and discrete turning actions from VSI-590K, and yaw changes between consecutive views and cumulative headings from SPAR-7M. We format each target type as a structured assistant response and train on these samples together with standard instruction data using the same autoregressive SFT objective.

\subsection{Joint Training Objective}
\label{sec:joint-training-objective}

\paragraph{Pairwise Geometric Prediction.}
Answer-level SFT constrains the final response but does not provide an explicit target for the relations encoded by the geometric branch. Because the geometric kernel operates on token pairs, we use the relative 3D displacement of ordered token pairs to directly supervise the geometric representation before fusion. We decompose the displacement into distance and unit direction to constrain its magnitude and orientation. At the layer $\ell^\star$ selected for pairwise geometric supervision, we read $\mathbf h_i=\phi_g^{\ell^\star}(G_i^{\ell^\star})\in\mathbb R^d$ after the shared output projection and before kernel fusion. We sample ordered pairs of visual tokens with valid geometry within and across frames, with $d_{ij}=\lVert\mathbf x_j-\mathbf x_i\rVert_2$. For each sampled pair, we construct the pair representation and its distance and direction targets as follows:

\begin{equation}
\begin{aligned}
\mathbf z_{ij}
&=[\mathbf h_i;\mathbf h_j;\mathbf h_j-\mathbf h_i;
\mathbf h_i\odot\mathbf h_j]
\in\mathbb R^{4d},\\
y^d_{ij}&=\log(1+d_{ij}),\qquad
\mathbf u_{ij}=\frac{\mathbf x_j-\mathbf x_i}{d_{ij}}
\quad(d_{ij}>\varepsilon).
\end{aligned}
\label{eq:pairwise-targets}
\end{equation}

Here, $\odot$ denotes elementwise multiplication. A prediction head maps $\mathbf z_{ij}$ to a predicted log-distance $\widehat y^d_{ij}$ and an L2-normalized direction $\widehat{\mathbf u}_{ij}$. For a sample $n$, we use a Huber loss on log distance, $\ell_{\mathrm{dist}}^{(n)}$, and a cosine loss on direction, $\ell_{\mathrm{dir}}^{(n)}$, combined as $\ell_{\mathrm{pair}}^{(n)}=\ell_{\mathrm{dist}}^{(n)}+0.5\,\ell_{\mathrm{dir}}^{(n)}$. Averaging over samples with valid token pairs gives $\mathcal L_{\mathrm{pair}}$; Appendix~\ref{app:pairwise-objective} specifies the exact reductions and empty-set cases.

\paragraph{Unified Objective.}
The two objectives are jointly optimized in a single training stage:

\begin{equation}
\mathcal L
=
\mathcal L_{\mathrm{SFT}}
+
\lambda(t)\mathcal L_{\mathrm{pair}},
\label{eq:joint-objective}
\end{equation}

where $\lambda(t)$ controls the pairwise-loss weight. The prediction head is used only during training and removed for inference.

\section{Experiments}
\label{sec:experiments}

% TODO before submission: reconcile reported/reproduced baseline sources and
% training settings across Tables 1, 2, and 4, including GeoThinker and VG-LLM.
% Keep this reminder out of the visible text and captions while results are updated.

\definecolor{tablebest}{RGB}{184,160,218}
\definecolor{tablesecond}{RGB}{229,220,241}

\begin{table}[t]
    \centering
    \caption{\textbf{Spatial reasoning on VSI-Bench and ReVSI-32.} Both panels report overall scores and task results. Higher is better. Darker and lighter shading mark the best and second best results in each column. $^*$ denotes results obtained using public checkpoints and official evaluation code.}
    \label{tab:vsibench}
    \scriptsize
    \setlength{\tabcolsep}{2.2pt}
    \renewcommand{\arraystretch}{1.06}
    \begin{tabular*}{\textwidth}{@{\extracolsep{\fill}}lccccccccc@{}}
        \toprule
        \multicolumn{10}{l}{\textbf{(a) VSI-Bench}} \\
        \textbf{Model} & \textbf{Overall} & \shortstack{\textbf{Obj.}\\\textbf{Count}} & \shortstack{\textbf{Abs.}\\\textbf{Dist.}}
        & \shortstack{\textbf{Obj.}\\\textbf{Size}} & \shortstack{\textbf{Room}\\\textbf{Size}} & \shortstack{\textbf{Rel.}\\\textbf{Dist.}}
        & \shortstack{\textbf{Rel.}\\\textbf{Dir.}} & \shortstack{\textbf{Route}\\\textbf{Plan}} & \shortstack{\textbf{Appr.}\\\textbf{Order}} \\
        \midrule
        Qwen2.5-VL-7B & 29.30 & 25.20 & 10.50 & 36.40 & 29.60 & 38.40 & 38.00 & 29.80 & 26.80 \\
        Qwen3-VL-8B & 57.90 & 67.60 & 47.00 & 76.30 & 61.90 & 58.00 & 51.00 & 35.00 & 66.30 \\
        \midrule
        VG-LLM-8B & 62.20 & 71.40 & 56.80 & 69.00 & 69.10 & 67.90 & 83.20 & 47.40 & 32.50 \\
        VLM-3R-7B & 60.90 & 70.20 & 49.40 & 69.20 & 67.10 & 65.40 & 80.50 & 45.40 & 40.10 \\
        Cambrian-S-7B & 67.50 & 73.20 & 50.50 & 74.90 & 72.20 & 71.10 & 76.20 & 41.80 & 80.10 \\
        SpatialStack-5B & 67.50 & 71.00 & 55.60 & 69.10 & 68.20 & 67.30 & 84.10 & 41.20 & \cellcolor{tablebest}\textbf{83.50} \\
        Q-GeoMem-7B & 71.20 & \cellcolor{tablesecond}74.80 & 59.90 & \cellcolor{tablesecond}76.50 & \cellcolor{tablesecond}75.30 & 71.10 & 87.20 & 50.00 & 74.60 \\
        GeoThinker-7B & 68.50 & 71.96$^*$ & 56.89$^*$ & 74.34$^*$ & 73.37$^*$ & 70.42$^*$ & 85.89$^*$ & 46.39$^*$ & 68.77$^*$ \\
        GeoThinker-8B & \cellcolor{tablesecond}72.60 & 73.93$^*$ & \cellcolor{tablebest}\textbf{62.49}$^*$ & 76.23$^*$ & \cellcolor{tablebest}\textbf{75.31}$^*$ & \cellcolor{tablesecond}72.96$^*$ & \cellcolor{tablesecond}90.29$^*$ & 49.48$^*$ & 76.21$^*$ \\
        \midrule
        DK-VLM-7B & 69.42 & 72.02 & 54.17 & 74.01 & 64.97 & 68.73 & 87.41 & \cellcolor{tablesecond}56.19 & 77.83 \\
        \textbf{DK-VLM-8B} & \cellcolor{tablebest}\textbf{74.33} & \cellcolor{tablebest}\textbf{75.91} & \cellcolor{tablesecond}60.48 & \cellcolor{tablebest}\textbf{77.82} & 72.78 & \cellcolor{tablebest}\textbf{74.79} & \cellcolor{tablebest}\textbf{90.48} & \cellcolor{tablebest}\textbf{60.31} & \cellcolor{tablesecond}82.04 \\
        \bottomrule
    \end{tabular*}

    \medskip
    \begin{tabular*}{\textwidth}{@{\extracolsep{\fill}}lcccccccc@{}}
        \toprule
        \multicolumn{9}{l}{\textbf{(b) ReVSI-32}} \\
        \textbf{Model} & \textbf{Overall} & \shortstack{\textbf{Obj.}\\\textbf{Count}} & \shortstack{\textbf{Abs.}\\\textbf{Dist.}}
        & \shortstack{\textbf{Obj.}\\\textbf{Size}} & \shortstack{\textbf{Room}\\\textbf{Size}} & \shortstack{\textbf{Rel.}\\\textbf{Dist.}}
        & \shortstack{\textbf{Rel.}\\\textbf{Dir.}} & \shortstack{\textbf{Route}\\\textbf{Plan}} \\
        \midrule
        Qwen3-VL-8B-Instruct & 45.60 & 36.20 & 47.10 & 67.20 & 45.30 & 50.20 & 37.50 & 36.00 \\
        InternVL3.5-38B & 53.10 & \cellcolor{tablesecond}43.40 & 56.50 & \cellcolor{tablesecond}70.00 & 56.90 & 55.80 & 47.80 & 41.50 \\
        \midrule
        VLM-3R-7B & 50.10 & 41.60 & 61.60 & 64.80 & 52.50 & 46.50 & \cellcolor{tablebest}\textbf{49.50} & 34.10 \\
        GeoThinker-7B & 48.96$^*$ & 37.66$^*$ & 65.22$^*$ & 62.16$^*$ & 50.28$^*$ & 38.74$^*$ & 47.15$^*$ & 41.47$^*$ \\
        GeoThinker-8B & \cellcolor{tablesecond}54.16$^*$ & 36.99$^*$ & \cellcolor{tablebest}\textbf{71.23}$^*$ & 65.67$^*$ & 53.32$^*$ & \cellcolor{tablesecond}61.35$^*$ & 47.54$^*$ & 43.02$^*$ \\
        \midrule
        DK-VLM-7B & 52.02 & \cellcolor{tablebest}\textbf{44.30} & 61.69 & 66.65 & \cellcolor{tablesecond}60.20 & 37.58 & 46.84 & \cellcolor{tablesecond}46.90 \\
        \textbf{DK-VLM-8B} & \cellcolor{tablebest}\textbf{58.28} & 42.23 & \cellcolor{tablesecond}70.55 & \cellcolor{tablebest}\textbf{72.70} & \cellcolor{tablebest}\textbf{60.99} & \cellcolor{tablebest}\textbf{65.51} & \cellcolor{tablesecond}48.35 & \cellcolor{tablebest}\textbf{47.67} \\
        \bottomrule
    \end{tabular*}
    \vspace{-1\baselineskip}
\end{table}

\subsection{Experimental Setup}
\label{sec:experimental-setup}

\paragraph{Implementation Details.}
We build DK-VLM on Qwen2.5-VL-7B and Qwen3-VL-8B~\citep{bai2025qwen25vl,bai2025qwen3vl}.
A frozen VGGT-$\Omega$ estimates 3D points and camera poses from the input images~\citep{wang2026vggtomega}.
We also evaluate alternative geometry frontends without retraining and report the results in Appendix~\ref{app:frontend-and-frames}.
We uniformly sample up to 32 frames from each training video.
For both models, we uniformly select five decoder layers to introduce Dual-Kernel Attention and apply rank-64 LoRA~\citep{hu2022lora} to the language model's Q/K/V/O projections.
The geometric projections, gates, and pairwise prediction head are also trained, while the visual encoder remains frozen.
For the 7B and 8B variants, the trainable parameter counts are only 2.88\% and 3.86\% of their respective backbone parameter counts.
Both models use AdamW~\citep{loshchilov2019adamw} with a main learning rate of $10^{-4}$ and a global batch size of 32.
Further implementation details are provided in Appendix~\ref{app:implementation-details}.

\paragraph{Training Data.}
We use the training data from GeoThinker's scaled regime~\citep{li2026geothinker} and retain all original instruction samples.
We construct approximately 1.6M RC samples with stepwise supervision from existing VSI-590K and SPAR annotations.
After filtering and sampling, we use 150K RC training samples containing only two- and three-step chains.
We train on the combined dataset for one epoch, jointly optimizing the SFT and pairwise geometric objectives.

\paragraph{Evaluation.}
We evaluate on seven spatial benchmarks: VSI-Bench, ReVSI-32, MMSI-Bench, MindCube, ViewSpatial, SITE, and CV-Bench~\citep{yang2025vsibench,zhang2026revsi,yang2025mmsibench,yin2025mindcube,li2025viewspatial,wang2025site,tong2024cambrian1}.
We follow GeoThinker's evaluation settings on the six shared benchmarks~\citep{li2026geothinker} and use the official 32-frame setting for ReVSI.
Our comparisons cover base VLMs~\citep{wang2025internvl35}, spatial post-training methods, and geometry-enhanced models~\citep{li2026geothinker,zhang2026spatialstack,gao2026qgeomem}.
We also evaluate RC on public tasks involving multiple reasoning steps or changes in reference frame, and assess general video understanding on VideoMME and MVBench~\citep{fu2025videomme,li2024mvbench}.

\subsection{Spatial Reasoning}
\label{sec:spatial-results}

\paragraph{VSI-Bench and ReVSI-32.}
DK-VLM-8B achieves the highest overall scores on both benchmarks, reaching 74.33 and 58.28, respectively (Table~\ref{tab:vsibench}).
On VSI-Bench, both the 7B and 8B variants outperform the corresponding GeoThinker models.
The 8B variant also outperforms GeoThinker-8B on route planning in both benchmarks, showing that the full method improves performance on tasks that require combining several spatial relations.

\begin{table}[t]
    \centering
    \caption{\textbf{Results across six spatial benchmarks.} Avg. is the mean of the six overall scores. Higher is better.}
    \label{tab:spatial-benchmarks}
    \scriptsize
    \setlength{\tabcolsep}{3pt}
    \renewcommand{\arraystretch}{1.08}
    \begin{tabular*}{\textwidth}{@{\extracolsep{\fill}}lccccccc@{}}
        \toprule
        \textbf{Model} & \textbf{Avg.} & \shortstack{\textbf{VSI-}\\\textbf{Bench}} & \shortstack{\textbf{MMSI-}\\\textbf{Bench}} & \textbf{MindCube} & \textbf{ViewSpatial} & \textbf{SITE} & \shortstack{\textbf{CV-}\\\textbf{Bench}} \\
        \midrule
        Qwen2.5-VL-7B & 40.31 & 29.30 & 26.80 & 36.00 & 36.80 & 37.60 & 75.40 \\
        Qwen3-VL-8B & 47.73 & 57.90 & 28.80 & 29.80 & 39.00 & 45.80 & \cellcolor{tablebest}\textbf{85.10} \\
        \midrule
        Cambrian-S-7B & 47.28 & 67.50 & 25.80 & 39.60 & 40.90 & 33.00 & 76.90 \\
        VLM-3R-7B & 45.40 & 60.90 & 27.90 & 40.00 & 40.50 & 31.30 & 71.80 \\
        VG-LLM-8B & 51.05 & 62.20 & 30.00 & 36.10 & 45.80 & 50.50 & 81.70 \\
        GeoThinker-7B & 60.43 & 68.50 & \cellcolor{tablebest}\textbf{31.70} & 83.60 & 41.90 & 54.80 & 82.10 \\
        GeoThinker-8B & \cellcolor{tablesecond}62.23 & \cellcolor{tablesecond}72.60 & \cellcolor{tablesecond}30.90 & 83.00 & \cellcolor{tablesecond}45.90 & 55.90 & \cellcolor{tablebest}\textbf{85.10} \\
        \midrule
        DK-VLM-7B & 60.84 & 69.42 & 29.95 & \cellcolor{tablesecond}83.81 & 44.98 & \cellcolor{tablesecond}57.47 & 79.38 \\
        \textbf{DK-VLM-8B} & \cellcolor{tablebest}\textbf{64.09} & \cellcolor{tablebest}\textbf{74.33} & 30.79 & \cellcolor{tablebest}\textbf{83.83} & \cellcolor{tablebest}\textbf{51.25} & \cellcolor{tablebest}\textbf{59.81} & \cellcolor{tablesecond}84.52 \\
        \bottomrule
    \end{tabular*}
    \vspace{-0.5\baselineskip}
\end{table}

\paragraph{Results across Spatial Benchmarks.}
To assess overall spatial understanding, we further compare models across six spatial benchmarks (Table~\ref{tab:spatial-benchmarks}).
DK-VLM-8B achieves the highest average score of 64.09, outperforming the spatial post-training and geometry-enhanced methods in the table.
It also ranks first on four of the six benchmarks, showing that the gains extend beyond a single benchmark.

\subsection{Effect of RC Supervision}
\label{sec:rc-results}

\begin{table}[t]
    \centering
    \caption{\textbf{Effect of RC supervision on public tasks.} Both models use Qwen3-VL-8B. Person denotes the average score across three person-perspective tasks in ViewSpatial: relative direction, object view orientation, and scene simulation relative direction.}
    \label{tab:rc-results}
    \scriptsize
    \setlength{\tabcolsep}{3pt}
    \renewcommand{\arraystretch}{1.1}
    \begin{tabular*}{\textwidth}{@{\extracolsep{\fill}}lcccc@{}}
        \toprule
        \textbf{Model} & \shortstack{\textbf{VSI}\\\textbf{Route}} & \shortstack{\textbf{ReVSI}\\\textbf{Route}} & \shortstack{\textbf{MMSI}\\\textbf{Multi-Step}} & \shortstack{\textbf{ViewSpatial}\\\textbf{Person}} \\
        \midrule
        \textbf{DK-VLM (Full)} & \textbf{60.31} & \textbf{47.67} & \textbf{34.34} & \textbf{61.72} \\
        DK-VLM w/o RC & 51.21 & 43.48 & 29.19 & 56.32 \\
        \bottomrule
    \end{tabular*}
    \vspace{-3pt}
\end{table}

To assess whether RC supervision improves performance on public tasks involving multiple reasoning steps or changes in reference frame, we compare the full DK-VLM-8B with a version trained without RC supervision (Table~\ref{tab:rc-results}).
Both retain Dual-Kernel Attention and pairwise geometric supervision.
RC improves route planning on VSI-Bench and ReVSI-32, as well as multi-step reasoning on MMSI-Bench.
These gains show that geometric relation modeling benefits from additional stepwise supervision, helping the model use spatial relations across reasoning steps.
Improvements on the person-perspective tasks in ViewSpatial further suggest that the benefits extend beyond sequential reasoning to spatial judgments under a different reference frame.

\noindent\begin{minipage}{\textwidth}
\begin{wraptable}{r}{0.44\textwidth}
    \vspace{-\intextsep}
    \setlength{\abovecaptionskip}{0pt}
    \centering
    \caption{\textbf{Performance across reasoning chain lengths.} Each entry reports Progress / Chain Acc. (\%). See Appendix~\ref{app:rc-length-evaluation} for metric definitions.}
    \label{tab:rc-length-results}
    \small
    \setlength{\tabcolsep}{8pt}
    \renewcommand{\arraystretch}{1.1}
    \begin{tabular}{@{}ccc@{}}
        \toprule
        \textbf{Chain Length} & \textbf{w/o RC} & \textbf{w/ RC} \\
        \midrule
        2 & 10.6 / 2.2 & \textbf{52.2 / 45.6} \\
        3 & 9.0 / 0.3 & \textbf{50.0 / 37.5} \\
        4 (unseen) & 8.2 / 0.0 & \textbf{35.4 / 20.6} \\
        \bottomrule
    \end{tabular}
\end{wraptable}

\paragraph{Performance across Chain Lengths.}
To assess whether the benefits of RC extend to chain lengths unseen during training, we evaluate stepwise spatial prediction on a held-out set constructed using our RC procedure, with four-step chains excluded from RC training.
Progress measures the fraction of steps correct before the first error; Chain Acc. measures full-sequence accuracy.
Table~\ref{tab:rc-length-results} shows that RC improves both metrics at every evaluated length.
The gains on unseen four-step chains suggest that RC helps the model apply spatial relations more consistently across reasoning steps.
\par
\ifnum\value{WF@wrappedlines}>1
  \dimen0=\baselineskip\multiply\dimen0 by \value{WF@wrappedlines}\vskip\dimen0
\fi
\WFclear
\end{minipage}\par

\subsection{General Video Understanding}
\label{sec:video-results}

\noindent\begin{minipage}{\textwidth}
\begin{wraptable}{r}{0.50\textwidth}
    \vspace{-\intextsep}
    \setlength{\abovecaptionskip}{0pt}
    \centering
    \caption{\textbf{General video understanding.} Parentheses show changes from the corresponding Qwen SFT baseline in percentage points.}
    \label{tab:video-results}
    \small
    \setlength{\tabcolsep}{3pt}
    \renewcommand{\arraystretch}{1.1}
    \begin{tabular*}{\linewidth}{@{\extracolsep{\fill}}lcc@{}}
        \toprule
        \textbf{Model} & \textbf{VideoMME} & \textbf{MVBench} \\
        \midrule
        Qwen3-VL-8B + SFT & 64.59 & \textbf{68.42} \\
        \textbf{DK-VLM-8B} & \textbf{66.22} (+1.63) & 67.93 ($-0.49$) \\
        \bottomrule
    \end{tabular*}
\end{wraptable}

We evaluate whether the gains in spatial reasoning are accompanied by a decline in general video understanding.
The baseline is Qwen3-VL-8B fine-tuned on the same data, including RC samples.
DK-VLM-8B scores 1.63 points higher on VideoMME and 0.49 points lower on MVBench (Table~\ref{tab:video-results}), indicating comparable general video performance.
\par
\ifnum\value{WF@wrappedlines}>1
  \dimen0=\baselineskip\multiply\dimen0 by \value{WF@wrappedlines}\vskip\dimen0
\fi
\WFclear
\end{minipage}\par

\subsection{Ablation Studies}
\label{sec:ablations}

We use Qwen3-VL-8B to examine the contributions of geometric transforms and pairwise supervision, as well as the combined effect of the architecture and RC supervision, on VSI-Bench.

\noindent\begin{minipage}{\textwidth}
\begin{wraptable}{r}{0.44\textwidth}
    \vspace{-\intextsep}
    \setlength{\abovecaptionskip}{0pt}
    \centering
    \caption{\textbf{Component ablations.}}
    \label{tab:ablations}
    \small
    \setlength{\tabcolsep}{4pt}
    \renewcommand{\arraystretch}{1.1}
    \begin{tabular*}{\linewidth}{@{\extracolsep{\fill}}lr@{}}
        \toprule
        \textbf{Configuration} & \textbf{Overall} \\
        \midrule
        Full & \textbf{74.33} \\
        w/o Point GTA & 71.81 \\
        w/o Camera GTA & 72.46 \\
        w/o Point \& Camera GTA & 70.02 \\
        w/o Pairwise Supervision & 73.20 \\
        \bottomrule
    \end{tabular*}
\end{wraptable}

\paragraph{Geometric Modeling Components.}
Table~\ref{tab:ablations} shows that using Point GTA and Camera GTA together outperforms using either alone.
Replacing both with identity transforms lowers the score by 4.31 percentage points, with the full geometric branch and its parameters retained, showing that the gains extend beyond added capacity.
Pairwise geometric supervision improves the model's ability to make spatial judgments.
\par
\ifnum\value{WF@wrappedlines}>1
  \dimen0=\baselineskip\multiply\dimen0 by \value{WF@wrappedlines}\vskip\dimen0
\fi
\WFclear
\end{minipage}\par\medskip

\noindent\begin{minipage}{\textwidth}
\begin{wraptable}{r}{0.44\textwidth}
    \vspace{-\intextsep}
    \setlength{\abovecaptionskip}{0pt}
    \centering
    \caption{\textbf{Effect of RC supervision.}}
    \label{tab:rc-ablation}
    \small
    \setlength{\tabcolsep}{3pt}
    \renewcommand{\arraystretch}{1.5}
    \begin{tabular*}{\linewidth}{@{\extracolsep{\fill}}lrrr@{}}
        \toprule
        \textbf{Model} & \textbf{w/o RC} & \textbf{w/ RC} & $\boldsymbol{\Delta}$ \\
        \midrule
        Backbone & 68.80 & 69.20 & +0.40 \\
        DK-VLM & 73.28 & \textbf{74.33} & +1.05 \\
        \bottomrule
    \end{tabular*}
\end{wraptable}

\paragraph{Architecture and RC Supervision.}
To examine whether geometric relation modeling and RC supervision are complementary, we train DK-VLM and Qwen3-VL-8B on the same instruction data, each with and without RC.
RC yields a larger gain on DK-VLM (Table~\ref{tab:rc-ablation}), supporting the complementarity between geometric relation modeling and stepwise supervision.
\par
\ifnum\value{WF@wrappedlines}>1
  \dimen0=\baselineskip\multiply\dimen0 by \value{WF@wrappedlines}\vskip\dimen0
\fi
\WFclear
\end{minipage}\par

\section{Conclusion}
\label{sec:conclusion}

We present DK-VLM, which introduces 3D geometry as position into the attention computation of vision-language models.
Dual-Kernel Attention preserves the native semantic kernel and uses Point GTA and Camera GTA to model relations between scene points and camera motion across views.
Reference-Frame Composition supervision reorganizes existing single-step annotations into training chains with stepwise targets, further improving spatial reasoning over multiple steps.
Among the compared methods, DK-VLM-8B achieves the highest average score across six spatial benchmarks while preserving general video understanding.
Ablations further support the complementarity between geometric relation modeling and stepwise supervision.

\section*{AI Use Statement}
We used AI assistants to help draft, translate, and polish the manuscript, retrieve literature, and implement code. The authors reviewed the AI-assisted content and code and take full responsibility for the accuracy and integrity of this work.

\bibliography{paper}

\appendix
\section{Additional Method Details}
\label{app:method-details}

\subsection{Geometry Preprocessing}
\label{app:geometry-preprocessing}

The geometric kernel of DK-VLM uses 3D points and camera poses aligned with visual tokens, without relying on intermediate features from a specific geometry frontend. For each merged visual token, we map the center of its image region to the frontend's dense 3D point map, accounting for resizing, cropping, and padding in both image preprocessing pipelines. Within each video, we filter out low-confidence points using the 20th percentile of finite depth confidence values as the threshold. We obtain each token's 3D coordinates by bilinear interpolation over the retained neighboring points, with the interpolation weights renormalized. All visual tokens from the same frame share that frame's camera pose.

We construct these geometric quantities from the frontend outputs and express them in the coordinate frame of the first camera. Within each video, we center and scale all point coordinates together and apply the same scale factor to camera translations. Point GTA uses the processed scene points as translation anchors and the source frame's rotation relative to the first frame. Camera GTA uses the relative camera poses. This common representation allows different geometry frontends to use the same geometric attention computation.

\subsection{Derivation of the Point GTA Relative Transform}
\label{app:point-gta-expansion}

In both DK-VLM-7B and DK-VLM-8B, each geometric attention head has 128 channels: 48 for Point GTA, 24 for Camera GTA, and 56 remaining channels. The point and camera subspaces contain 12 and 6 four-dimensional blocks, respectively. For the 12 Point GTA blocks, we use geometrically spaced displacement scales from 0.25 to 4.0, given by $\lambda_b=0.25\cdot16^{b/11}$ for $b=0,\ldots,11$.

For the Point GTA transform in Section~\ref{sec:dual-kernel},

\begin{equation}
T^{\mathrm{pt}}_{i,b}
=
\begin{bmatrix}
R_i^{\mathrm{rel}} & \lambda_b\mathbf x_i\\
\mathbf 0^{\top} & 1
\end{bmatrix},
\end{equation}

its inverse is

\begin{equation}
\bigl(T^{\mathrm{pt}}_{i,b}\bigr)^{-1}
=
\begin{bmatrix}
{R_i^{\mathrm{rel}}}^{\top} &
-\lambda_b{R_i^{\mathrm{rel}}}^{\top}\mathbf x_i\\
\mathbf 0^{\top} & 1
\end{bmatrix}.
\end{equation}

Multiplication with the transform of token $j$ gives the complete pairwise relation

\begin{equation}
\bigl(T^{\mathrm{pt}}_{i,b}\bigr)^{-1}
T^{\mathrm{pt}}_{j,b}
=
\begin{bmatrix}
{R_i^{\mathrm{rel}}}^{\top}R_j^{\mathrm{rel}} &
\lambda_b{R_i^{\mathrm{rel}}}^{\top}
(\mathbf x_j-\mathbf x_i)\\
\mathbf 0^{\top} & 1
\end{bmatrix}.
\label{eq:point-gta-expanded-relation}
\end{equation}

The upper-left and upper-right blocks are the relative source-view rotation $R_{ij}^{\mathrm{view}}$ and scaled point displacement $\mathbf t_{ij,b}^{\mathrm{pt}}$ retained in the main text.

\subsection{Pairwise Loss Reduction and Boundary Cases}
\label{app:pairwise-objective}

\paragraph{Prediction Head and Pair Sampling.}
Both models use LayerNorm followed by a two-layer MLP with GELU activation to map the $4d$-dimensional pair features to a log-distance prediction and a three-dimensional direction prediction, which is L2-normalized.
The MLP hidden dimension is $d$, set to 3584 and 4096 for the 7B and 8B variants, respectively.

We use a sampling budget of 256 ordered token pairs per sample, targeting 50\% cross-frame pairs for multi-frame inputs.
Within-frame and cross-frame candidates are sampled separately by distance tertiles, with unused quotas reassigned to the other group; single-frame inputs use only within-frame pairs.
The direction loss includes only pairs with distances greater than $\varepsilon=10^{-6}$.

For sample $n$, let $\mathcal J_n$ denote the indices of geometry-valid visual tokens. The sampled ordered-pair set and its direction-valid subset are

\begin{equation}
\begin{aligned}
\mathcal P_n
&\subseteq
\{(i,j):i,j\in\mathcal J_n,\ i\neq j\},\\
\mathcal V_n
&=
\{(i,j)\in\mathcal P_n:d_{ij}>\varepsilon\}.
\end{aligned}
\end{equation}

For each sample with $|\mathcal P_n|>0$, the distance and direction terms are

\begin{equation}
\begin{aligned}
\ell_{\mathrm{dist}}^{(n)}
&=
\frac{1}{|\mathcal P_n|}
\sum_{(i,j)\in\mathcal P_n}
\operatorname{Huber}_{\delta=1}
(\widehat y^d_{ij},y^d_{ij}),\\
\ell_{\mathrm{dir}}^{(n)}
&=
\begin{cases}
\displaystyle
\frac{1}{|\mathcal V_n|}
\sum_{(i,j)\in\mathcal V_n}
\left(1-\widehat{\mathbf u}_{ij}^{\top}\mathbf u_{ij}\right),
& |\mathcal V_n|>0,\\[8pt]
0, & |\mathcal V_n|=0,
\end{cases}\\
\ell_{\mathrm{pair}}^{(n)}
&=\ell_{\mathrm{dist}}^{(n)}+0.5\,\ell_{\mathrm{dir}}^{(n)}.
\end{aligned}
\end{equation}

Let $\mathcal B_+=\{n:|\mathcal P_n|>0\}$. The pairwise loss over the minibatch, including the case with no valid pairs, is

\begin{equation}
\mathcal L_{\mathrm{pair}}
=
\begin{cases}
\displaystyle
\frac{1}{|\mathcal B_+|}
\sum_{n\in\mathcal B_+}\ell_{\mathrm{pair}}^{(n)},
& |\mathcal B_+|>0,\\[8pt]
0, & |\mathcal B_+|=0.
\end{cases}
\label{eq:pairwise-full-reduction}
\end{equation}

Pairs with $d_{ij}\leq\varepsilon$, including zero-distance pairs, remain in the distance term but do not enter the direction denominator. When $\mathcal V_n=\varnothing$, the direction loss is zero but remains connected to the predicted directions in the computation graph. Samples with $\mathcal P_n=\varnothing$ do not enter the pairwise batch average; when $\mathcal B_+=\varnothing$, the batch pairwise loss is zero. Otherwise, samples with valid pairs contribute equally to $\mathcal L_{\mathrm{pair}}$.

\subsection{Reference-Frame Composition Details}
\label{app:reference-frame-composition}

Reference-Frame Composition uses planar angular relations extracted from the source annotations. We write $\langle z\rangle_{360}=z\bmod 360^\circ$ for normalization to $[0,360^\circ)$, with clockwise rotation positive, and use the corresponding circular distance $d_\circ$ for tolerance checks. We first parse and normalize the source relations, then apply consistency checks specific to each source to obtain the verified single-step annotations used for chain construction. After these checks, scenes are partitioned, and chains are constructed only within one scene and split. During partitioning, scene groups that overlap VSI-Bench are isolated from the RC pool. Finally, an independent validator rereads the source annotations, recomputes every derived target, and verifies scene and split membership together with source provenance. These checks cross-validate source angle annotations and their algebraic consistency; they do not use images, XYZ coordinates, or ground-truth 3D states.

\paragraph{VSI-RC.}
From each VSI-590K \texttt{absolute\_direction\_object} annotation~\citep{yang2025cambrians}, the parser extracts

\begin{equation}
(\mathrm{media},O,A,B,\mathrm{convention},\alpha(O,A,B)),
\end{equation}

where $O$ is the current location, $A$ is the facing anchor, and $B$ is the target landmark. A clockwise raw angle is normalized directly, whereas a counterclockwise angle is converted as $\langle 360^\circ-a_{\mathrm{raw}}\rangle_{360}$. The normalized angle must agree with the corresponding \texttt{relative\_direction\_object} label under the source four-quadrant convention. Inverse relations and angle closure are checked with a $2^\circ$ circular tolerance:

\begin{equation}
\begin{aligned}
d_\circ\!\left(\left\langle\alpha(O,A,B)+\alpha(O,B,A)\right\rangle_{360},0\right)&\leq2^\circ,\\
d_\circ\!\left(\left\langle\alpha(O,A,C)-\alpha(O,A,B)\right\rangle_{360},\alpha(O,B,C)\right)&\leq2^\circ.
\end{aligned}
\end{equation}

Verified relations are connected only within the same media item. A relation extends the current path when its origin is the terminal landmark and its facing anchor is the preceding landmark; paths with repeated landmarks are rejected. After the initial turn, a landmark path $\mathcal C=(v_0,\ldots,v_K)$ induces the ordered reference-state transitions $(v_{i-1},v_i)\rightarrow(v_i,v_{i+1})$ for $1\leq i<K$. Let $a_0$ be the initial facing anchor. At subsequent steps, the source annotation uses the direction toward the previous landmark as its reference. This reference is opposite to the direction of arrival, requiring a $180^\circ$ correction. Converting the source facing convention to the traversal-centered convention gives

\begin{equation}
\delta_k^{\mathrm{VSI}}
=
\begin{cases}
\alpha(v_0,a_0,v_1), & k=1,\\
\left\langle\alpha(v_{k-1},v_{k-2},v_k)-180^\circ\right\rangle_{360}, & 2\leq k\leq K.
\end{cases}
\end{equation}

The exact-turn target retains the full sequence $(\delta_1^{\mathrm{VSI}},\ldots,\delta_K^{\mathrm{VSI}})$. The discrete-action target is created only when every turn lies in one of the conservative bins: straight for $[0^\circ,30^\circ]\cup[330^\circ,360^\circ)$, right for $[60^\circ,120^\circ]$, back for $[150^\circ,210^\circ]$, and left for $[240^\circ,300^\circ]$. VSI-RC does not serialize cumulative headings. Its prompt specifies the initial reference point, the initial bearing target, and the ordered waypoints while withholding the derived turn sequence.

\paragraph{SPAR-RC.}
From each SPAR-7M \texttt{view\_change\_infer} annotation with decodable structured camera-motion fields~\citep{zhang2025spar}, the parser extracts the source and target view identifiers and the yaw component. Yaw is converted to the same clockwise-positive angular convention; pitch and translation fields are not used in RC. Repeated records for the same directed view pair are deduplicated. For three available relations $A\!\rightarrow\!B$, $B\!\rightarrow\!C$, and $A\!\rightarrow\!C$, triangle closure requires

\begin{equation}
d_\circ\!\left(\theta_{AC},\left\langle\theta_{AB}+\theta_{BC}\right\rangle_{360}\right)\leq5^\circ.
\end{equation}

Accepted directed relations are connected when the target view of one edge matches the source view of the next; paths with repeated views are rejected. For transition $v_{k-1}\rightarrow v_k$, $\delta_k^{\mathrm{SPAR}}=\theta_{v_{k-1}v_k}$; cumulative headings follow the recurrence in Section~\ref{sec:reference-frame-composition}. SPAR-RC applies no facing-to-traversal $180^\circ$ correction. Its prompt provides the ordered views and their images and uses the first view as the heading origin while withholding the target sequence.

\paragraph{Supervision Targets.}
RC produces four target types: VSI \texttt{angle} contains exact turn sequences; VSI \texttt{action} contains the corresponding discrete actions for eligible chains; SPAR \texttt{delta} contains local yaw increments; and SPAR \texttt{heading} contains cumulative headings relative to the initial view. Each target sequence is serialized as a JSON array in the assistant response of a separate training sample. No sample jointly emits two target types. The validator independently reconstructs these arrays and checks their provenance against the source relations before they enter autoregressive SFT.

\subsection{Output Projection and Gate Initialization}
\label{app:gate-initialization}

At each selected layer, we initialize the shared output projection $\phi_g^{\ell}$ by copying the native output projection. The two projections have separate parameters. At positions with valid geometry, both kernel outputs use $\phi_g^{\ell}$; at other positions, the semantic output uses the native output projection.

We initialize $g_\ell=0$, giving both kernels an initial mixing weight of 0.5. This avoids strongly suppressing the geometric branch at the start of training and keeps the sigmoid gate away from saturation. Each selected layer then learns its own mixing ratio.

\subsection{Implementation Details and Parameter Counts}
\label{app:implementation-details}

We introduce Dual-Kernel Attention at layers 4, 9, 14, 19, and 24 for DK-VLM-7B and at layers 6, 12, 18, 27, and 33 for DK-VLM-8B, using zero-based indices. The pairwise prediction head reads the pre-fusion geometric features at layer 19 for the 7B model and layer 27 for the 8B model. For both models, the pairwise loss weight increases linearly to 0.1 over the first 5\% of training steps.

Both models are trained on 32 NVIDIA H200 GPUs for one epoch. Training takes approximately 11 days for the 7B variant and 10 days for the 8B variant.

Table~\ref{tab:parameter-counts} summarizes the parameter counts of both variants.

\begin{table}[!htbp]
    \centering
    \caption{\textbf{Parameter counts of DK-VLM.} Parameter counts are reported in millions (M). Trainable parameters include LoRA, the geometric branch, and the pairwise prediction head. The prediction head is used only during training and removed at inference. All counts exclude the frozen VGGT-$\Omega$ frontend.}
    \label{tab:parameter-counts}
    \small
    \setlength{\tabcolsep}{12pt}
    \renewcommand{\arraystretch}{1.1}
    \begin{tabular}{@{}lrr@{}}
        \toprule
        \textbf{Component} & \textbf{DK-VLM-7B} & \textbf{DK-VLM-8B} \\
        \midrule
        Backbone & 8,292.17 & 8,767.12 \\
        LoRA & 40.37 & 61.34 \\
        Geometric branch (including gates) & 146.82 & 209.72 \\
        Pairwise prediction head & 51.43 & 67.16 \\
        \midrule
        \textbf{Total trainable parameters} & \textbf{238.62} & \textbf{338.22} \\
        \textbf{Relative to backbone} & \textbf{2.88\%} & \textbf{3.86\%} \\
        \bottomrule
    \end{tabular}
\end{table}

\subsection{Training Data Composition}
\label{app:training-data}

We follow the training mixture reported for GeoThinker's scaled regime~\citep{li2026geothinker}, which contains approximately 1.8M instruction samples. The spatial reasoning data comprise Cambrian-S VSI-590K (590K), SPAR (234K), VLM-3R VSI-Bench instructions (205K), VLM-3R VSTI-Bench instructions (132K), and the MindCube training set (10K). The general video data comprise LLaVA-Hound (64K), PhysGame PhysInstruct (140K), and a general video subset of Cambrian-S-3M (430K).

\paragraph{RC Sample Selection.}
We construct a pool of approximately 1.6M RC samples from the verified source annotations. For SPAR-RC, we retain one question per view tuple and randomly assign either local yaw increments or cumulative headings as the prediction target. We also retain at most one sample for each image set.

To reduce the concentration of small turns, we group individual turn magnitudes into six $30^\circ$ bins over the folded angular range from $0^\circ$ to $180^\circ$. Each chain is represented by the multiset of its step-level bins. We use deficit-greedy sampling to favor underrepresented bins, subject to limits of 96 samples per scene and 40 appearances per image.

For VSI-RC, we retain only samples with video inputs and limit each clip to 40 questions. The retained VSI-RC samples are repeated twice to increase their weight in the training mixture. The final RC training set contains 150K entries and includes only two- and three-step chains.

\subsection{RC Chain Evaluation}
\label{app:rc-length-evaluation}

\paragraph{Evaluation Set Construction.}
We construct an evaluation set using the same RC procedure, with no overlap with the training samples. It includes four target types: local yaw increments and cumulative headings from SPAR-RC, and turn angles and discrete actions from VSI-RC. We evaluate two-, three-, and four-step chains. RC training uses only two- and three-step chains.

For each chain length, we select up to 80 examples per target type. To cover different angular configurations, SPAR-RC examples are grouped by data source (ScanNet, ScanNet++, or S3D) and the $30^\circ$ bin containing the largest folded turn magnitude in the chain. VSI-RC angle examples are grouped by the same angular criterion, while action examples are grouped by their complete action sequences. Examples are selected in round-robin order across groups within each target type.

This gives 320 examples each for two- and three-step chains. Only eight distinct four-step action examples are available, giving 248 examples for four-step chains. The composition of the evaluation set is shown in Table~\ref{tab:rc-eval-composition}.

\begin{table}[!htbp]
    \centering
    \caption{\textbf{Composition of the RC evaluation set.} Counts denote evaluation examples for each target type and chain length.}
    \label{tab:rc-eval-composition}
    \small
    \setlength{\tabcolsep}{5pt}
    \renewcommand{\arraystretch}{1.1}
    \begin{tabular*}{\textwidth}{@{\extracolsep{\fill}}lccccc@{}}
        \toprule
        \textbf{Chain Length} & \textbf{SPAR Delta} & \textbf{SPAR Heading} & \textbf{VSI Angle} & \textbf{VSI Action} & \textbf{Total} \\
        \midrule
        2 & 80 & 80 & 80 & 80 & 320 \\
        3 & 80 & 80 & 80 & 80 & 320 \\
        4 & 80 & 80 & 80 & 8 & 248 \\
        \bottomrule
    \end{tabular*}
\end{table}

\paragraph{Evaluation Metrics.}
Both models are evaluated on the same examples using the same correctness criteria. For numeric angle targets, a step is correct if the shortest circular difference between the predicted and target angles is at most $10^\circ$; discrete actions require exact label matching. We report Progress and Chain Acc. separately for each chain length.

For a target sequence of length $n$, let $k$ be the number of consecutive correct predictions from the first step. Progress is $100k/n$. Once a prediction is incorrect, that step and all subsequent steps receive no credit, even if a later prediction matches its target. For example, a three-step prediction with only its first step correct receives a Progress score of $33.3\%$.

Chain Acc. assigns a score of $100$ only when the predicted sequence has the same length as the target and every step is correct, and $0$ otherwise. Both metrics are averaged over all evaluation examples of the corresponding chain length, pooling the four target types.

\clearpage
\subsection{Injection Layer Configuration}
\label{app:injection-layer-configuration}

We examine the number and placement of Dual-Kernel Attention layers in DK-VLM-8B. To reduce training cost, all configurations are trained on the same 10\% subset of the full training data, with the remaining training settings held constant. All layer indices are zero-based. Pairwise geometric supervision is applied at the penultimate selected injection layer in each configuration.

\paragraph{Number of Injection Layers.}
Increasing the number of injection layers from three to five improves VSI-Bench performance by 0.77 percentage points, while increasing it from five to seven yields only a further 0.06-point gain (Table~\ref{tab:injection-layer-count}). We therefore use five injection layers by default, achieving performance close to the seven-layer configuration with fewer geometric branches.

\begin{table}[!htbp]
    \centering
    \caption{\textbf{Effect of the number of injection layers on VSI-Bench.}}
    \label{tab:injection-layer-count}
    \small
    \setlength{\tabcolsep}{12pt}
    \renewcommand{\arraystretch}{1.1}
    \begin{tabular}{@{}clc@{}}
        \toprule
        \textbf{Injection Layers} & \textbf{Layer Indices} & \textbf{VSI-Bench} \\
        \midrule
        3 & 6, 18, 33 & 62.28 \\
        5 & 6, 12, 18, 27, 33 & 63.05 \\
        7 & 6, 9, 12, 18, 24, 27, 33 & \textbf{63.11} \\
        \bottomrule
    \end{tabular}
\end{table}

\paragraph{Placement of Injection Layers.}
With the number of injection layers fixed at five, we compare early, distributed, and late placements. The distributed configuration achieves the highest score among the three, outperforming the early and late configurations by 0.62 and 1.76 percentage points, respectively (Table~\ref{tab:injection-layer-placement}). These results support our default choice of distributing geometric injection across model depth.

\begin{table}[!htbp]
    \centering
    \caption{\textbf{Effect of injection layer placement on VSI-Bench.} All configurations use five injection layers.}
    \label{tab:injection-layer-placement}
    \small
    \setlength{\tabcolsep}{12pt}
    \renewcommand{\arraystretch}{1.1}
    \begin{tabular}{@{}llc@{}}
        \toprule
        \textbf{Placement} & \textbf{Layer Indices} & \textbf{VSI-Bench} \\
        \midrule
        Early & 6, 9, 12, 15, 18 & 62.43 \\
        Distributed & 6, 12, 18, 27, 33 & \textbf{63.05} \\
        Late & 18, 21, 24, 27, 30 & 61.29 \\
        \bottomrule
    \end{tabular}
\end{table}

\clearpage
\subsection{Geometry Frontends and Input Frame Counts}
\label{app:frontend-and-frames}

We examine the sensitivity of DK-VLM-8B to the geometry frontend and the number of input frames on VSI-Bench. Both experiments use the same trained checkpoint, changing only the inference configuration without additional training. We fix the input to 32 frames when comparing frontends and use VGGT-$\Omega$ throughout the frame-count comparison.

\paragraph{Geometry Frontends.}
DK-VLM uses estimated 3D points and camera poses rather than frontend-specific intermediate features. To test whether this interface supports interchangeable frontends, we replace the default VGGT-$\Omega$ with other geometry frontends at inference. As shown in Table~\ref{tab:geometry-frontends}, VGGT-$\Omega$ achieves the highest score of 74.33, while the other frontends score between 72.99 and 73.91, a decrease of 0.42--1.34 percentage points from the default setting. These results show that DK-VLM can use different geometry frontends without retraining while maintaining spatial reasoning performance close to that of the default configuration.

\begin{table}[!htbp]
    \centering
    \caption{\textbf{Geometry frontend comparison on VSI-Bench.} All configurations use the same DK-VLM-8B checkpoint and 32 input frames, without retraining.}
    \label{tab:geometry-frontends}
    \small
    \setlength{\tabcolsep}{20pt}
    \renewcommand{\arraystretch}{1.1}
    \begin{tabular}{@{}lc@{}}
        \toprule
        \textbf{Geometry Frontend} & \textbf{Overall} \\
        \midrule
        VGGT-$\Omega$ & \textbf{74.33} \\
        VGGT & 73.91 \\
        MASt3R-SLAM & 73.64 \\
        Pi3X & 73.62 \\
        DA3-Large & 72.99 \\
        \bottomrule
    \end{tabular}
\end{table}

\paragraph{Input Frame Counts.}
With VGGT-$\Omega$ fixed, we increase the number of input frames at inference from 8 to 128. Performance improves with more frames, but the gains diminish: increasing from 8 to 32 frames yields a 9.04-point improvement, whereas increasing from 32 to 128 frames adds only 0.99 points (Table~\ref{tab:input-frame-count}). The 128-frame setting achieves the highest score of 75.32; our default 32-frame setting approaches this performance with fewer input frames.

\begin{table}[!htbp]
    \centering
    \caption{\textbf{Effect of input frame count on VSI-Bench.} All configurations use the same DK-VLM-8B checkpoint and VGGT-$\Omega$ frontend, without additional training.}
    \label{tab:input-frame-count}
    \small
    \setlength{\tabcolsep}{20pt}
    \renewcommand{\arraystretch}{1.1}
    \begin{tabular}{@{}cc@{}}
        \toprule
        \textbf{Input Frames} & \textbf{Overall} \\
        \midrule
        8 & 65.29 \\
        16 & 69.94 \\
        32 & 74.33 \\
        64 & 75.13 \\
        128 & \textbf{75.32} \\
        \bottomrule
    \end{tabular}
\end{table}

\subsection{General Video Evaluation Settings}
\label{app:video-evaluation}

We use up to 32 frames per video for both VideoMME and MVBench, and evaluate VideoMME without subtitles.

\clearpage
\subsection{Comparison of Geometry Injection Strategies}
\label{app:geometry-injection}

We compare different geometry injection strategies based on Qwen3-VL-8B using the same 10\% training data subset and optimization settings. All configurations omit the pairwise geometric loss. In addition to Qwen SFT and DK-VLM, we consider three alternatives: adding 3D positional embeddings to visual tokens once at the input; adding 3D positional embeddings to visual tokens at the same five selected decoder layers as DK-VLM, while retaining single-stream attention at each layer; and applying single-kernel GTA at three or five selected decoder layers, directly replacing the positional encoding in the corresponding channels of the native attention. DK-VLM scores 62.78 in this comparison; the score of 63.05 reported in the injection layer experiments (Appendix~\ref{app:injection-layer-configuration}) corresponds to the five-layer configuration with the pairwise geometric loss retained.

\begin{table}[!htbp]
    \centering
    \caption{\textbf{Comparison of geometry injection strategies on VSI-Bench.} All configurations omit the pairwise geometric loss.}
    \label{tab:geometry-injection}
    \small
    \setlength{\tabcolsep}{12pt}
    \renewcommand{\arraystretch}{1.1}
    \begin{tabular}{@{}lr@{}}
        \toprule
        \textbf{Configuration} & \textbf{VSI-Bench Overall} $\uparrow$ \\
        \midrule
        Qwen3-VL-8B + SFT & 59.48 \\
        3D positional embeddings at input & 59.21 \\
        3D positional embeddings at five layers & 59.74 \\
        Single-kernel GTA (three layers) & 45.98 \\
        Single-kernel GTA (five layers) & 39.14 \\
        \textbf{DK-VLM-8B (five layers)} & \textbf{62.78} \\
        \bottomrule
    \end{tabular}
\end{table}

Adding 3D positional embeddings at the input and at five layers yields scores of 59.21 and 59.74, respectively, close to the Qwen SFT score of 59.48. In contrast, DK-VLM without the pairwise geometric loss still achieves 62.78, outperforming Qwen SFT and five-layer embedding addition by 3.30 and 3.04 points, respectively (Table~\ref{tab:geometry-injection}). This shows that, under the current training settings, dual-kernel geometry injection outperforms the two tested additive positional embedding strategies.

Single-kernel GTA scores 45.98 with three injection layers and 39.14 with five, both below the Qwen SFT baseline. With five injection layers, DK-VLM outperforms single-kernel GTA by 23.64 points. In the five-layer single-kernel configuration, the performance drop is accompanied by garbled outputs and poorer instruction following, including responses that do not follow the required answer format. These observations suggest that directly replacing the native positional encoding can impair model performance and output quality in this setting. Overall, these results provide empirical support for our design of retaining the native attention pathway while introducing spatial relations through a separate geometric kernel.

\end{document}